\documentclass[10pt,twocolumn,letterpaper]{article}

\usepackage{iccv}
\usepackage[table]{xcolor}
\usepackage{times}
\usepackage{epsfig}
\usepackage{graphicx}
\usepackage{amsmath}
\usepackage{colortbl}
\usepackage{amssymb}
\usepackage{booktabs}
\usepackage{color}
\usepackage{float}
\usepackage{array}
\usepackage{multirow}
\usepackage{makecell}
\usepackage{amssymb}
\usepackage{caption}
\usepackage{soul}
\usepackage{xcolor}
\usepackage{multirow}
\usepackage{cuted}
\usepackage{capt-of}
\usepackage[OT1]{fontenc}
\usepackage{paralist}
\usepackage{bm}
\usepackage[super]{nth}
\usepackage[pagebackref=true,colorlinks,bookmarks=false,citecolor=blue,linkcolor=blue,urlcolor=blue]{hyperref}
\newcommand{\jk}[1]{\textcolor{blue}{#1}}

\newcommand{\ql}[1]{\textcolor{purple}{#1}}

\definecolor{LightCyan}{rgb}{0.5,1,1}

\usepackage{comment}
\usepackage[capitalize]{cleveref}
\crefname{section}{Sec.}{Secs.}
\Crefname{section}{Section}{Sections}
\Crefname{table}{Table}{Tables}
\crefname{table}{Tab.}{Tabs.}

\def\eg{\emph{e.g}.}

\def\etc{\emph{etc}}

\iccvfinalcopy 

\def\iccvPaperID{x} 

\ificcvfinal\fi

\begin{document}

\title{High-Quality Entity Segmentation}

\author{
Lu Qi$^{1}$\thanks{Equal contribution.},
~~~
Jason Kuen$^{2\ast}$,~~~
Weidong Guo$^{3\ast}$,~~~
Tiancheng Shen$^{4\ast}$,~~~
Jiuxiang Gu$^2$,~~~ \\
Jiaya Jia$^4$, ~~~
Zhe Lin$^2$, ~~~
Ming-Hsuan Yang$^1$, ~~~
\\[0.2cm]
$^1$The University of California, Merced~~
$^2$Adobe Research~~ \\
$^3$QQ Brower Lab, Tencent,~~
$^4$The Chinese University of Hong Kong~~
}

\maketitle
\ificcvfinal\thispagestyle{empty}\fi

\begin{strip}
  \centering
  \includegraphics[height=0.5\linewidth, width=0.99\linewidth]{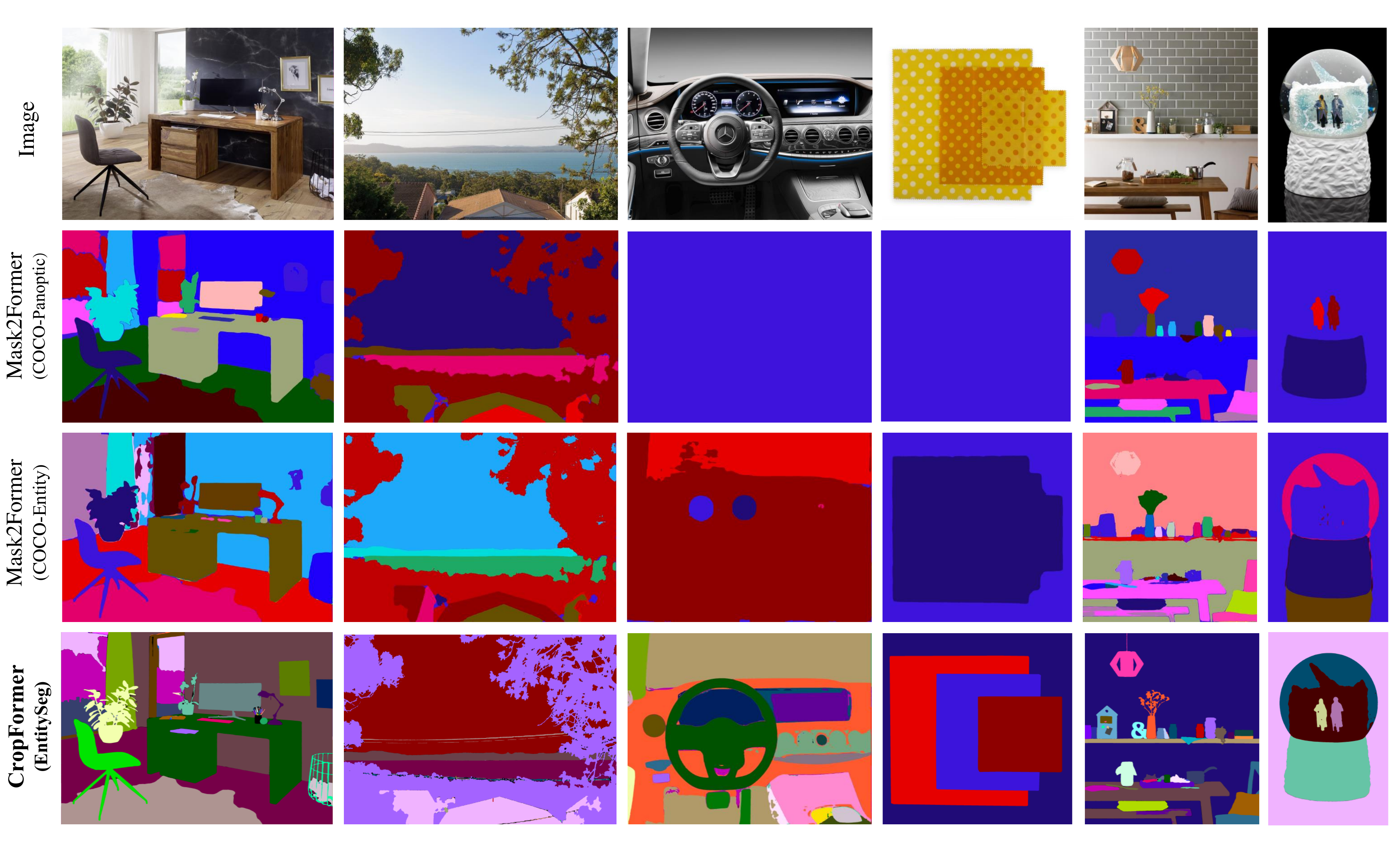}
\captionof{figure}{Segmentation results on in-the-wild test images: (1) Mask2Former~\cite{cheng2022masked} 
trained for panoptic segmentation~\cite{kirillov2019panoptic} on COCO dataset~\cite{lin2014microsoft}
; (2) Mask2Former trained for entity segmentation~\cite{qi2021open} on COCO dataset~\cite{lin2014microsoft}; (3) our CropFormer 
trained on the proposed EntitySeg Dataset. For a fair comparison, all three models use Swin-Large backbone~\cite{liu2021swin} and are trained until full convergence. Our approach provides far more desirable and useful results for many real-world applications. 
}
\label{fig:comp_coco}
\end{strip}

\begin{abstract}
Dense image segmentation tasks (\textit{e.g.,} semantic, panoptic) are useful for image editing, but existing methods can hardly generalize well in an in-the-wild setting where there are unrestricted image domains, classes, and image resolution \& quality variations.
Motivated by these observations, we construct a new entity segmentation dataset, with a strong focus on high-quality dense segmentation in the wild. 
The dataset contains images spanning diverse image domains and entities, along with plentiful high-resolution images and high-quality mask annotations for training and testing. 
Given the high-quality and -resolution nature of the dataset, we propose CropFormer which is designed to tackle the intractability of instance-level segmentation on high-resolution images.
It improves mask prediction by fusing high-res image crops that provides more fine-grained image details and the full image. CropFormer is the first query-based Transformer architecture that can effectively fuse mask predictions from multiple image views, by learning queries that effectively associate the same entities across the full image and its crop. With CropFormer, we achieve a significant AP gain of $1.9$ on the challenging entity segmentation task. Furthermore, CropFormer consistently improves the accuracy of traditional segmentation tasks and datasets. The dataset and code will be released at \href{http://luqi.info/entityv2.github.io/}{http://luqi.info/entityv2.github.io/}.
\end{abstract}

\section{Introduction}
Entity segmentation~\cite{qi2021open} is an emerging segmentation task that focuses on open-world and class-agnostic dense image segmentation, designed to have superior generalization capabilities for unseen-category segmentation. As such, it has great potential in image editing applications, where users can interact with visual entities that exist ``in the wild''. Recent works have leveraged entity segmentation for image manipulation \cite{wang2022manitrans, wang2023entity} and image retouching \cite{liu2022text}, where users usually specify in-the-wild text queries (\textit{e.g., Oculus, platypus}) for local editing. In comparison to traditional closed-set category-based segmentation tasks such as panoptic/instance/semantic segmentation, entity segmentation has the advantage of being all-inclusive and unrestricted, making it easier to find masks that semantically match well with user-specified queries.

Previous work on entity segmentation~\cite{qi2021open} has shown promising results, yet it only employed existing panoptic segmentation datasets to verify the feasibility and benefits of entity segmentation. However, existing panoptic segmentation datasets (\textit {e.g.,} COCO~\cite {lin2014microsoft}, Cityscapes~\cite {cordts2016cityscapes}, ADE20K~\cite {zhou2017scene}) suffer from the same issues of closed-set categories that entity segmentation~\cite {qi2021open} aims to avoid. These datasets are collected and annotated for their original panoptic segmentation task requirements. In view of the lack of a proper entity segmentation dataset, it is difficult to further study and explore on this important task. To address this issue, we present EntitySeg dataset, a new large-scale and high-quality dataset for entity segmentation.

Apart from in-the-wild generalization, we design EntitySeg to prioritize high-quality mask annotations and high-resolution images, which are essential for user-driven image editing in the current era of high-resolution imagery. Image editing workflows usually require pixel-precise segmentation masks to ensure a smooth user experience. However, existing panoptic segmentation datasets generally contain low-resolution images (e.g., COCO\cite{lin2014microsoft} has zero images that are at least 1K) and coarsely-annotated masks, making it difficult for models trained on such datasets to produce accurate mask predictions for higher resolution images. This is why we focus on high-resolution images and high-quality masks in the EntitySeg dataset, which clearly distinguishes it from existing datasets.
The EntitySeg dataset poses new challenges to existing segmentation methods due to its high-resolution images and high-quality mask annotations. Standard practices of resizing images and mask annotations to 800px used in existing methods \cite{ren2015faster,he2017mask,cheng2022masked}, result in a heavy loss of fine-grained details. On the other hand, training or evaluating existing segmentation models directly on our high-resolution/quality dataset is intractable due to excessive computation and memory overheads. A naive workaround is to divide the high-resolution image into smaller patches and apply segmentation model to them separately, but it is unable to exploit global context to resolve ambiguity. Moreover, it is unclear how to combine mask predictions from multiple patches using state-of-the-art Transformer-based segmentation methods which rely on set-prediction \cite{cheng2022masked, cheng2021mask2former, wang2021max, yu2022k}, as shown in Fig~\ref{fig:framework} and our supplementary material.
%

To this end, we introduce CropFormer, a novel approach for tackling the problem of high-resolution entity segmentation. CropFormer learns to dynamically generate tokens to associate the full image with its high-resolution crops and then fuses them to generate the final mask prediction. During training, one of the predefined four corner crops is randomly selected and fused with the full image. For inference, CropFormer is applied to all four corner crops and the full image to achieve complete fusion. The strength of CropFormer lies in its ability to dynamically generate queries to link the same entities in the full image and its higher-resolution crops, allowing us to benefit from both global image context and high-quality image/annotation details to obtain improved results.

The main contributions of this work are as follows:
\begin{compactitem}
    \item We present a large-scale, high-quality entity segmentation dataset, comprising of images collected from various domains with diverse resolutions and high-quality masks. Our dataset allows for the evaluation of the generalization and robustness of segmentation models in a unified and open-world manner, due to its domain-diverse and high-resolution setting.
    
    \item We benchmark existing segmentation models on our newly collected dataset. We find that the traditional training setting cannot accommodate our dataset, which comprises over 70\% images with very high resolution.

    \item We propose a novel model, CropFormer, with an innovative association module and batch-level decoder to enable the fusion of the full image and its higher-resolution crops.
    
    \item Extensive experiments and thorough analysis demonstrate the efficacy of the proposed model and dataset on various segmentation tasks and settings. In Fig.~\ref{fig:comp_coco}, we compare the results from our best CropFormer (trained on EntitySeg) with the best Mask2Formers \cite{cheng2022masked} trained on COCO-Panoptic \cite{kirillov2019panoptic} and COCO-Entity \cite{qi2021open}.
\end{compactitem}

\section{Related Work}

\begin{figure*}[t!]
  \centering
  \includegraphics[width=.95\linewidth]{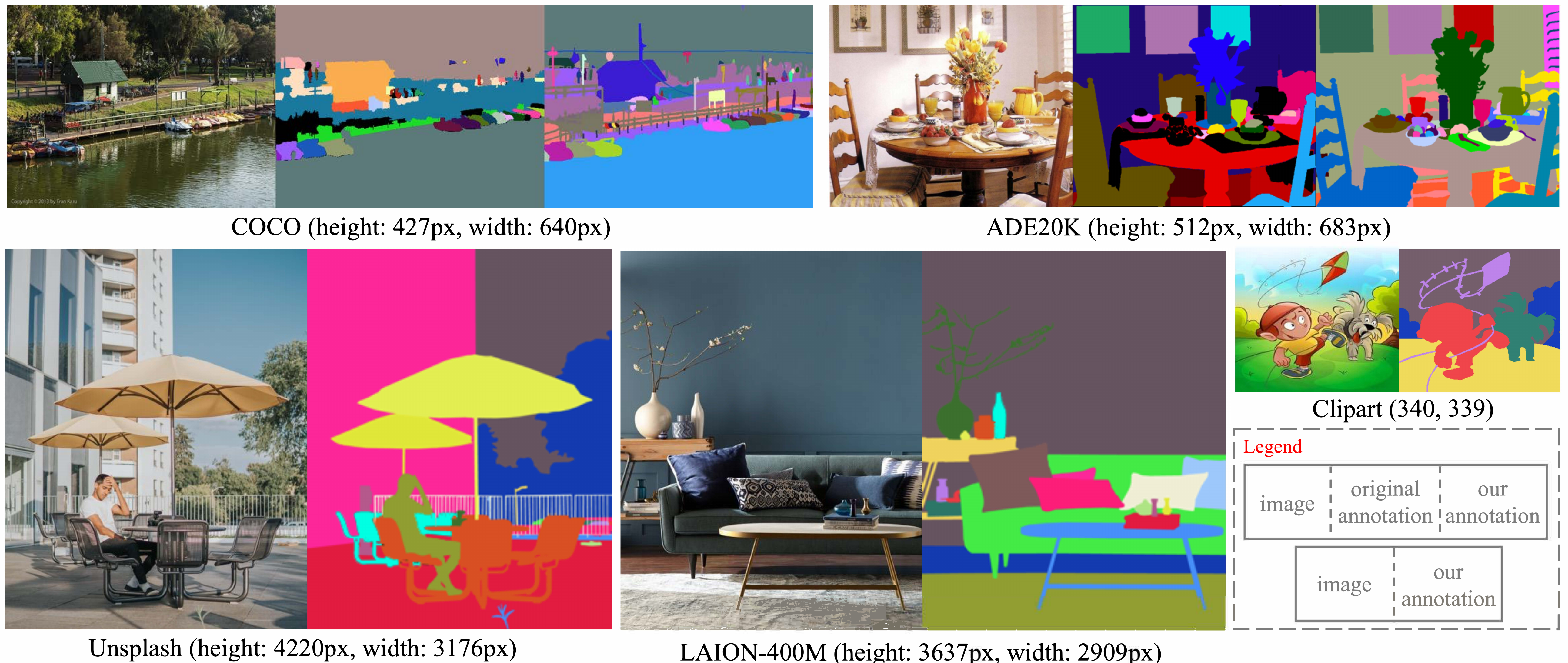}
  \caption{High-quality mask annotations for low- and high-resolution images collected from existing datasets such as COCO~\cite{lin2014microsoft}, ADE20K~\cite{zhou2017scene} and Cityscapes~\cite{cordts2016cityscapes} as well as from the internet are presented. For the images collected from the aforementioned datasets, a visual comparison between the original and our annotations is provided in the middle and rightmost sub-figures, where the unannotated regions are shaded in black. It is worth noting that the RGB and mask images shown here have been downsampled, resulting in some quality degradation compared to the original datasets.}
  \label{fig:dataset_vis}
\end{figure*}

\paragraph{Image Segmentation Dataset.}
Numerous datasets have been proposed for semantic, instance, and panoptic segmentation, \eg, Microsoft COCO~\cite{lin2014microsoft}, ADE20K~\cite{zhou2017scene}, KITTI~\cite{geiger2012we}, LVIS~\cite{gupta2019lvis}, PASCAL-VOC~\cite{everingham2010pascal}, Open-Images~\cite{kuznetsova2020open}, Cityscapes~\cite{cordts2016cityscapes}, and so on. In addition, there are some datasets specially designed for specific scenarios, such as amodal segmentation (COCO-Amodal~\cite{zhu2017semantic}, KINS~\cite{qi2019amodal}), human segmentation (CelebAMask-HQ~\cite{CelebAMask-HQ}, LIP~\cite{gong2017look}, MHP~\cite{zhao2018understanding}) and domain adaptation (Synscapes~\cite{wrenninge2018synscapes}, GTA5~\cite{richter2016playing}). Despite the significant contributions from these datasets, there is still a need to fulfill the requirements of real-world applications with high-quality images of large diversity.
%
For example, a segmentation model should be robust to high-resolution images from different domains and able to segment unseen objects in the open-world setting. Most relevant to our work is the ADE20K dataset~\cite{zhou2017scene}, which has large-scale open-vocabulary categories. However, the collected images are of low resolution and from narrowed domains. In contrast, our dataset collects images from multiple domains, including indoor, outdoor, street scenes, cartoon, and remote images, with over 80\% of the resolution of the images falling within the high-resolution range of 2000px to 8000px. Differently from the ADE20K~\cite{zhou2017places} and LVIS~\cite{gupta2019lvis} dataset with a predefined category list, our annotation process first conducts class-agnostic mask annotation on each entity and then labels the category information in an open-vocabulary manner.

\paragraph{Scene Parsing Methods.}

Scene parsing methods mainly rely on convolution-based dense prediction or transformer-based query prediction. Explicit localization information, such as bounding box proposals or pixel position, is often used for pixel-level mask prediction in these approaches. Examples include FCN~\cite{long2015fully}, DeepLab~\cite{chen2017deeplab}, PSPNet~\cite{zhao2017pyramid}, Mask R-CNN~\cite{he2017mask}, PANet~\cite{liu2018path}, SOLO~\cite{wang2019solo}, CondInst~\cite{tian2020conditional}, PanopticFPN~\cite{kirillov2019panopticfpn}, and PanopticFCN~\cite{li2022fully}. Inspired by the success of transformer-based detection tasks such as DETR~\cite{carion2020end}~\cite{lin2017feature,lin2017focal,dai2017deformable,qi2021multi}, a number of methods, such as Max-Deeplab~\cite{wang2021max} and Mask2Former~\cite{cheng2022masked}, have been proposed to directly predict segmentation masks without relying on bounding boxes. The key to these methods is to use learnable queries trained by a set-prediction loss to model the implicit location of the potential instance area. However, query-based set-prediction methods cannot effectively perform multi-view fusion, as the query-to-object assignment can drastically change with the change of image view. This is because the set prediction loss typically adopted by such methods does not enforce any assignment constraints across multiple image views.

\section{EntitySeg Dataset}

The EntitySeg dataset contains 33,227 images with high-quality mask annotations, as illustrated in Fig.~\ref{fig:dataset_vis}. Compared with existing datasets, it has three distinctive features. Firstly, 71.25\% and 86.23\% of the images are of high quality, with at least 2000px$\times$2000px and 1000px$\times$1000px, respectively, staying in line with the current digital imaging trends. Secondly, our dataset is open-world and not restricted to predefined classes. We consider each semantically-coherent region in the images as an entity, even if it is blurred or hard to recognize semantically. Thirdly, the mask annotation along the boundaries is more precise than those in other datasets. In the following, we will provide an in-depth description of our EntitySeg dataset, as well as an extensive analysis.

\begin{figure*}[t!]
  \centering
  \includegraphics[width=0.95\linewidth]{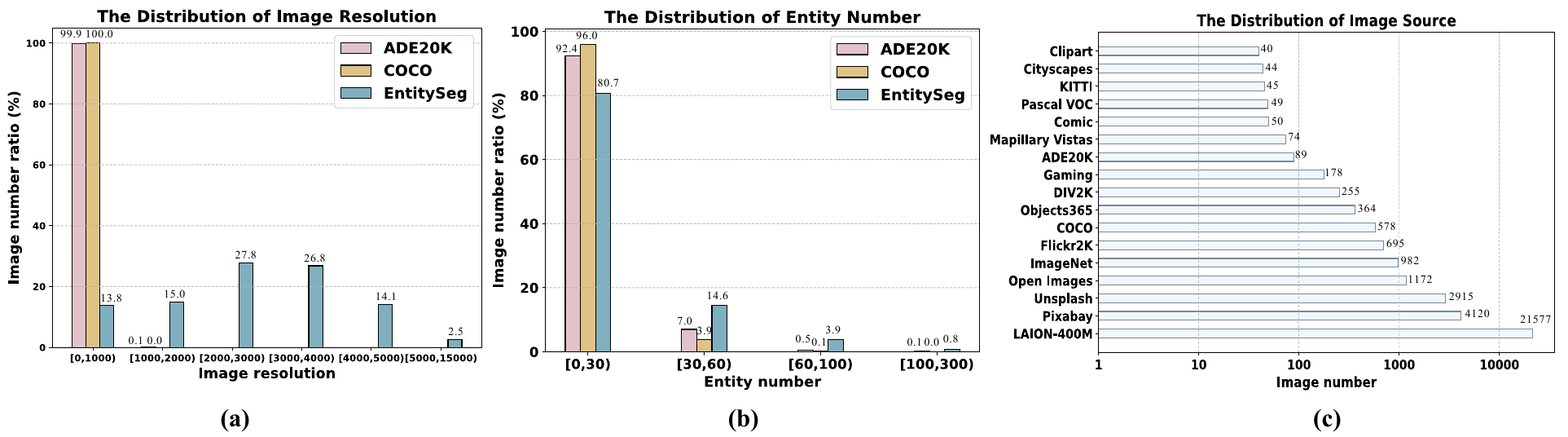}
  \caption{Sub-figures (a) and (b) present the distributions of image resolutions and average number of entities among ADE20K, COCO, and EntitySeg, respectively. Moreover, sub-figure (c) displays the distribution of image sources from which the EntitySeg images were collected.}
  \label{fig:statistics_01}
\end{figure*}

\subsection{Image Collection}

Inspired by the collection criterion of COCO~\cite{lin2014microsoft}, we have collected most non-iconic images with plentiful contextual information and non-canonical viewpoints. To ensure substantial domain diversity, we have sourced the images from various public datasets and the Internet, with permission for academic research. Specifically, part of the images are from COCO~\cite{lin2014microsoft}, ADE20K~\cite{zhou2017scene}, Pascal VOC~\cite{everingham2010pascal}, Cityscapes~\cite{cordts2016cityscapes}, Mapillary Vistas~\cite{neuhold2017mapillary}, ImageNet~\cite{krizhevsky2017imagenet}, Open Images~\cite{kuznetsova2020open}, DIV2K~\cite{agustsson2017ntire}, Flick2K~\cite{wang2019flickr1024}, Clipart~\cite{inoue2018cross}, Comic~\cite{inoue2018cross}, DOTA~\cite{xia2018dota}, Synscapes~\cite{wrenninge2018synscapes} and GTA5~\cite{richter2016playing}. Moreover, we have crawled high-resolution images from Pixabay~\cite{pixabay}, Unsplash~\cite{Unsplash}, and LAION-400M~\cite{schuhmann2021laion}\footnote{We removed not-safe-for-work images via strict CLIP filtering.}. 
As shown in Fig.~\ref{fig:statistics_01}(c), most of the images are from high-resolution sources like Pixabay, Unsplash, and LAION-400M. Moreover, Fig.~\ref{fig:statistics_01}(a) reveals that our dataset has more high-resolution images, with resolutions normally distributed within the range of 0px to 15,000px, in comparison to ADE20K~\cite{zhou2017scene} and COCO~\cite{lin2014microsoft} with images all under 1000px.

\subsection{Image Annotation}
For our dataset, the annotation process consists of the following steps. Firstly, entities in the image are annotated with class-agnostic pixel-level masks, ensuring that no overlaps exist between masks. Subsequently, each mask is annotated with a class label in an open vocabulary manner. Two special considerations should be taken into account here:
1)  we aonotate the entity as `\textit{unknown}' or `\textit{blurred}' if it cannot be easily named or is severely blurred; 2) the entity is annotated as a `\textit{supercategory}\_other' if we only know it is a \textit{supercategory} but unable to determine its fine-grained class. Our annotation process is the opposite of those of the popular segmentation datasets like COCO~\cite{lin2014microsoft} and ADE20K~\cite{zhou2017places}, which first predefine their class sets before annotating masks based on the predefined classes.

Table~\ref{Tab:comp_entity_other} indicates that our dataset has a more considerable proportion of covered image areas and more masks on average compared to COCO- and ADE20K-Panoptic. This is attributed to our novel annotation process that enables us to take all semantically-coherent regions into account and assign class labels accordingly. Besides, our annotation procedure is more similar to the human visual system. As evidenced in~\cite{Marr1982Vision}, the human vision system is intrinsically class-agnostic and can recognize entities without comprehending their usage and purpose.

\begin{table}[t!]
\begin{minipage}{\textwidth}
\begin{minipage}[t]{0.21\textwidth}
            \centering
            \scriptsize
            \setlength{\tabcolsep}{2pt}
            \begin{tabular}{c|ccc}
             & \cellcolor{lightgray!30} ann$_{1}$ & \cellcolor{lightgray!30} ann$_{2}$ & \cellcolor{lightgray!30} ann$_{3}$\\ \hline
            \cellcolor{lightgray!30} ann$_{1}$ & - & - & -\\ 
            \cellcolor{lightgray!30} ann$_{2}$ & 90.6 & - & -\\ 
            \cellcolor{lightgray!30} ann$_{3}$ & 91.2 & 92.1 & -\\ \hline
            \end{tabular}
        
        (a)
        \end{minipage}
        \begin{minipage}[t]{0.30\textwidth}
        \centering
        \scriptsize
        \setlength{\tabcolsep}{2pt}
        \begin{tabular}{c|ccc}
             & \cellcolor{lightgray!30} ann$_{1}$ & \cellcolor{lightgray!30} ann$_{2}$ & \cellcolor{lightgray!30} ann$_{3}$\\ \hline
            \cellcolor{lightgray!30} ann$_{1}$ & - & - & -\\
            \cellcolor{lightgray!30} ann$_{2}$ & 95.4 & - & -\\ 
            \cellcolor{lightgray!30} ann$_{3}$ & 94.8 & 95.2 & -\\ \hline
        \end{tabular}
        
    (b)
        \end{minipage}

\end{minipage}
\caption{Annotation consistency of class-agnostic localization and class-aware categories among annotators.}
\label{tab:consistency}
\end{table}

%
Annotation consistency is crucial to any human-labeled dataset since it implies whether the annotation task is well-defined and proper. To study this, we randomly selected 500 images from EntitySeg ($1.5\%$ of the entire dataset), and asked another two annotators to annotate them after four months since the first round. Table~\ref{tab:consistency}(a) shows that the class-agnostic mask mAP in the first step of the annotation process compared to those of the other two annotators. We use one as ground truth and another one as prediction results. In Table~\ref{tab:consistency}(b), we also evaluate the category consistency by accuracy (ACC) under the same mask annotations. These two tables demonstrate that our EntitySeg has high annotation consistency in the mask and category labeling stages.

\subsection{Dataset Statistics}
\begin{table}[t!]
\centering
\tiny
\begin{tabular}{c|c|c|c|c|c|c}
\cellcolor{lightgray!30} Dataset & \cellcolor{lightgray!30} \makecell{ImageRes\\(avg)$\uparrow$} & \cellcolor{lightgray!30} \cellcolor{lightgray!30} \makecell{EntityNum\\(avg)$\uparrow$} & \cellcolor{lightgray!30} \makecell{EntityNum\\(max)$\uparrow$} & \cellcolor{lightgray!30} \makecell{Entity\\Complexity$\downarrow$} &
\cellcolor{lightgray!30} \makecell{Entity\\Simplicity$\downarrow$} &\cellcolor{lightgray!30} \makecell{Valid\\Area$\uparrow$}  \\ \hline
COCO & 522.5 & 11.2 & 95 & 0.758 & 0.581 & 0.891 \\
ADE20K & 461.3& 13.6 & \textbf{255} & 0.802 & 0.606 & 0.914 \\
EntitySeg & \textbf{2700.7} & \textbf{18.1} & 236 & \textbf{0.719} & \textbf{0.538} & \textbf{0.999} \\ \hline
\end{tabular}
\caption{The statistical comparisons between COCO~\cite{lin2014microsoft}, ADE20K~\cite{zhou2017scene} and EntitySeg. `ImageRes (avg)', `EntityNum (avg)', `Valid Area', `Entity Complexity', and `Entity Simplicity' refer to the average value of image resolution size, entity numbers per image, valid area ratio per image, average entity complexity and simplicity respectively. `EntityNum (max)' means the maximum per-image number of entities within each dataset.}
\label{Tab:comp_entity_other}
\end{table}

In our EntitySeg dataset, we annotate all images at the entity level, regardless of whether they belong to thing or stuff in previous datasets. As shown in Fig.~\ref{fig:comp_coco}, we show some low- and high-resolution photos and their mask annotations to highlight our high-quality pixel-level annotation quality.
For example, one can zoom in on the Unsplash image to see how high-quality the railing's mask is.

Table~\ref{Tab:comp_entity_other} shows the quantitative comparison among our dataset, COCO-Panoptic~\cite{lin2014microsoft}, and ADE20K-Panoptic~\cite{zhou2017scene}.
In the EntitySeg dataset, each image has 18.1 entities on average, which is more than 11.2 and 13.6 entities in COCO and ADE20K. For a fair comparison, we treat each uncountable stuff region as one segment although it is disconnected. The detailed comparison among those three datasets on the distribution of entity number is shown in Fig.~\ref{fig:statistics_01}(b). Furthermore, the shapes of entities in our dataset are more complex than those COCO~\cite{lin2014microsoft} and ADE20K~\cite{zhou2017scene} as represented by the columns `Entity Complexity' and `Entity Simplicity' of Table~\ref{Tab:comp_entity_other} where an entity with a large convexity and simplicity value means it is a simple shape (and both metrics achieve their maximum value of 1.0 for a circle~\cite{zhu2017semantic}). 
More details about how complexity and simplicity are calculated can be found in the supplementary material.

\paragraph{Class-Aware Annotation.}
We select a subset of 11,580 images from the dataset for class labeling, for investigating the diversity of semantic classes in our dataset. In our class annotation process, the class is labeled in an open-vocabulary and free-form manner, then the class set is continually expanded. After the class annotation process, we organize and merge the annotated classes according to Word-Tree~\cite{ILSVRC15}.
This results in 535 things and 109 stuff classes for those 11,580 images that we use to construct the class-aware EntityClass dataset. Please refer to our supplementary file for more analysis on EntityClass dataset.

\section{CropFormer}\label{sec:cropformer}
\begin{figure*}[t!] 
  \centering
   \includegraphics[width=1.0\linewidth]{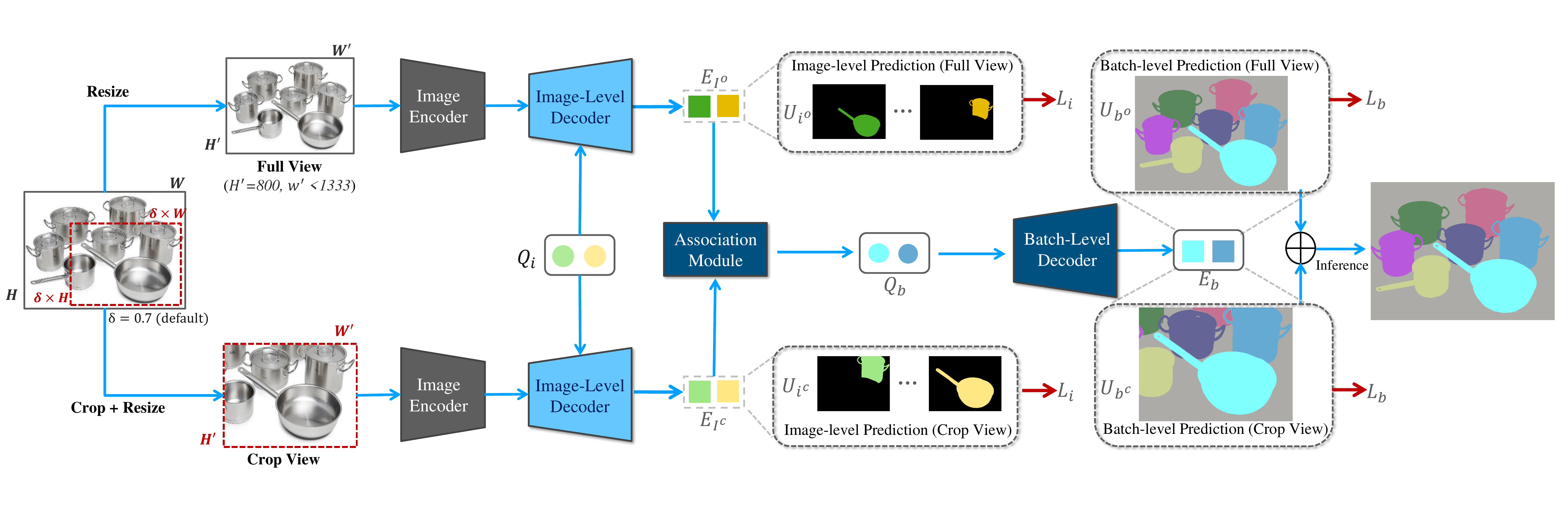}
   \caption{Framework of the proposed CropFormer. The red box indicates cropped region randomly sampled from four fixed image corners. In image-level prediction, the same entity across different image views may be assigned to different queries. Our association module and batch decoder can effectively associate the same entity across different views with a single query.}
   \label{fig:framework}
\end{figure*}

We propose CropFormer which is the first Transformer-based set-prediction segmentation method that is capable of fusing the mask predictions from multiple image views. Inspired by existing methods \cite{cheng2021per,cheng2022masked,wang2021max,yu2022k}, we design CropFormer to learn $N$ number of $K$-dimensional queries $\mathbf{Q}\in \mathbb{R}^{N\times K}$ to generate mask embeddings $\mathbf{E} \in \mathbb{R}^{N \times 1 \times 1 \times 1 \times K}$.  $\mathbf{E}$ then acts as convolution filters that are applied to pixel-level mask features $\mathbf{P_2} \in \mathbb{R}^{T \times H \times W}$ to generate $N$ segmentation masks $\mathbf{U^m} \in \mathbb{R}^{N \times T \times H \times W}$, where $T$, $H$ and $W$ correspond to the image view, height, and width dimensions. However, such a rudimentary design can only work well with single-view inputs and cannot effectively fuse the results from multiple views. Here, we introduce a novel association module and batch-level decoder to achieve the goal of CropFormer, which is to exploit global context from full image and fine-grained local details from crops for high-quality segmentation. CropFormer is illustrated in Fig.~\ref{fig:framework}.

\paragraph{Crop Dataloader.} In CropFormer framework, the crop dataloader is designed to generate a batch of images that simultaneously include the full images and their corresponding crops. There are two steps in our dataloader: \textit{crop} and \textit{resize}. First, we augment the full image $I^o$ with a crop that is part of the full image. The crop is randomly extracted from one of the fixed image corners: upper-left, upper-right, bottom-left, bottom-right. The crop size is controlled by a fixed ratio hyperparameter $\delta \in \mathbb{R}$ relative to the full image size. Then, we resize both the full image and crop to the same size. In this way, we construct an input tensor $\mathbf{I} \in \mathbb{R}^{1 \times 2 \times H' \times W' \times 3}$ with 2 elements (full image $\&$ 1 crop) along the view dimension. Compared to the full image, our corner crops preserve more image details (\eg, local information) that are useful for high-quality segmentation. Specifically, given an dataloader $\mathcal{D}$, we define the data preprocessing as:
\begin{equation}
\mathbf{I} = \{I^o, I^c \}, \ \ I^o \sim \mathcal{D}
\end{equation}
where $I^o$ is sampled from $\mathcal{D}$, and $I^c$ is the cropped image controlled by fixed ratio $\delta$. Unlike the random cropping in~\cite{seqformer,heo2022vita}, we adopt a fixed cropping solution to keep the training and inference consistent. Such a rigid cropping strategy provides better inference efficiency since, rather than bunches of random crops, evaluating a fixed number of crops during inference is sufficient enough.

\paragraph{Image Encoder and Decoder.} Based on learnable image-level queries $\mathbf{Q}_{i} \in \mathbb{R}^{N\times K}$,  we use image encoder $\Theta$ and image-level decoder $\Phi_{i}$ to generate image-level embeddings $\mathbf{E_i} \in \mathbb{R}^{N \times 2 \times 1 \times 1 \times K}$ for the full input image and its crop. 
Given the input tensor ($\mathbf{I}$) and queries ($\mathbf{Q}_{i}$), we define $\mathbf{E_i}$ as follows:
\begin{equation}
\mathbf{E_i} = \Phi_i(\mathbf{Q_i}, \Theta(\mathbf{I})),
\end{equation}
where $\Phi_{i}(\cdot)$ is a Transformer-based image-level decoder.
$\mathbf{E_i}$ is used for image-level entityness prediction and pixel-level prediction using the low-level image feature $\mathbf{P_2}$ (derived from the image encoder).
The prediction process can be formulated as:
\begin{equation}
\mathbf{U^e_i}, \mathbf{U^m_i} = \text{PredHead}_{\mathbf{i}}(\mathbf{E_i}, 
\mathbf{P_2})\label{eq:eq3},
\end{equation}
where $\mathbf{U^e_i}$ and $\mathbf{U^m_i}$ denote
the entityness prediction and pixel-level mask outputs. Here, we employ $i$ subscript to differentiate image-level embeddings and mask outputs from those of the association module.

\begin{figure}[t!] 
  \centering
   \includegraphics[height=1.3in, width=0.95\linewidth]{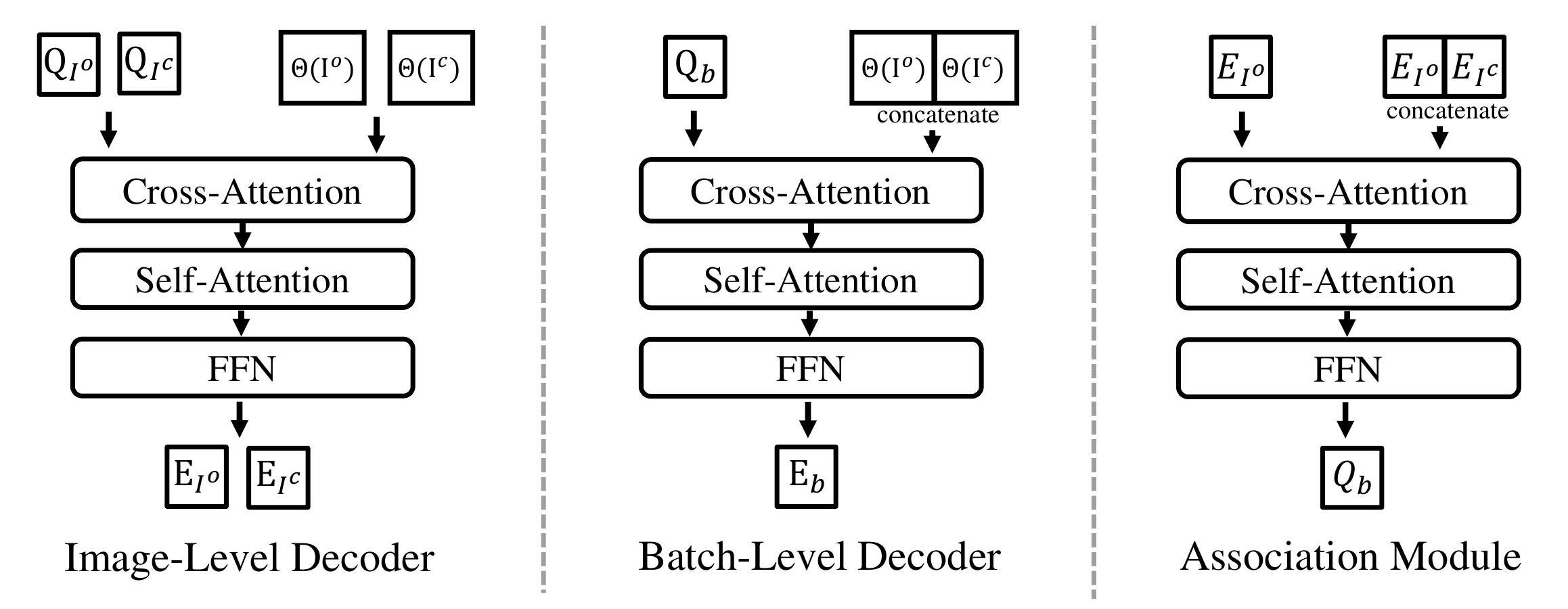}
   \caption{The illustration of image- and batch-level decoder and association module.}
   \label{fig:com_decoder}
\end{figure}

\paragraph{Association Module.}
Following~\cite{cheng2021mask2former}, our association module aims to generate batch queries $\mathbf{Q_b}$ that are fully shared by the full image and its crop to represent the same entities consistently. However, instead of treating $\mathbf{Q_b}$ as learnable network parameters, considering that $\mathbf{E_i}$ already contains strong segmentation features, we generate $\mathbf{Q_b}$ directly from $\mathbf{E_i}=\{\mathbf{E}_{I^o}, \mathbf{E}_{I^c}\}$.
Particularly, we use a Transformer architecture with cross-attention ($f_{\text{XAtt}}$) and self-attention ($f_{\text{SAtt}}$) to obtain $\mathbf{Q_b}$:
\begin{equation}
\mathbf{Q_{b}} = \text{FFN}(f_{\text{SAtt}}(f_{\text{XAtt}}(\underbrace{f_\text{q}(\mathbf{E}_{I^o})}_\text{query}, \underbrace{f_\text{k}(\mathbf{E_i})}_\text{key}, \underbrace{f_\text{v}(\mathbf{E_i}}_\text{value})))),
\end{equation}
where FFN is a feed-forward network,
$f_{\{\text{q},\text{k},\text{v}\}}(\cdot)$ are linear transformations.
Considering the importance of full image for entity segmentation, inspired by~\cite{vaswani2017attention,li2021selfdoc}
we take the image-level embeddings of full image $\mathbf{E}_{I^o}$ as \textit{query}, while treating all image-level embeddings $\mathbf{E_i}$ as \textit{key} and \textit{value}.
\paragraph{Batch-level Decoder.}
Given $\mathbf{Q_b} \in \mathbb{R}^{N \times 1 \times 1 \times 1 \times K}$, we obtain batch embeddings $\mathbf{E_b} \in \mathbb{R}^{N \times 1 \times 1 \times 1 \times K}$:
\begin{equation}
\mathbf{E_b} = \Phi_b(\mathbf{Q_b}, \Theta(\mathbf{I})),
\end{equation}
where $\Phi_b(\cdot)$ denotes the batch-level decoder. Thus, we broadcast $\mathbf{E_b}$ to the shape of $N \times 2 \times 1 \times 1 \times K$ to share it between the full image and its crop.
Finally, the broadcasted $\mathbf{E_b}$ and low-level batch features $\mathbf{P_{2}}$ are used for batch-level mask classification and pixel-wise prediction:
\begin{equation}
\mathbf{U^e_b}, \mathbf{U^m_b} = \text{PredHead}_{\mathbf{b}}(\mathbf{E_b}, 
\mathbf{P_{2}}),
\end{equation}
For the image-level decoder, the queries are separate for each image/crop. Whereas, for the batch-level decoder, the queries are shared between the image and its crop. We illustrate the differences of the image- and batch-level decoders in Fig.~\ref{fig:com_decoder}, where we regard two mage views as an 
entirety.

\paragraph{Training.} 
During training, we employ two separate losses $\mathcal{L}_{i}$ and $\mathcal{L}_{b}$ for image- and batch-level predictions, with respect to ground truth $\mathbf{G}$.
The main difference between the two losses lies in
whether the same entities in full image and crop are tied to the same queries or not.

The overall training loss for our model is defined as:
\begin{align}
\mathcal{L} = & \sum_{\mathbf{k}\in \{\mathbf{i}, \mathbf{b}\}} \mathcal{L}_{\mathbf{k}}^{\text{ce}}(\mathbf{U^e_k}, \mathbf{G^e_k}) + \sum_{{\mathbf{k}}\in \{\mathbf{i}, \mathbf{b}\}} \mathcal{L}_{\mathbf{k}}^{\text{bce}}(\mathbf{U^m_k}, \mathbf{G^m_k}) + \nonumber\\
& \sum_{\mathbf{k}\in \{\mathbf{i}, \mathbf{b}\}} \mathcal{L}_{\mathbf{k}}^{\text{dice}}(\mathbf{U^m_k}, \mathbf{G^m_k}),
\end{align}
where $\mathcal{L}_{\mathbf{i}}^{\text{ce}}$ and $\mathcal{L}_{\mathbf{b}}^{\text{ce}}$ denote binary cross-entropy loss for image- and batch-level entityness prediction. Similarly, $\mathcal{L}_{\mathbf{i}}^{\text{bce}}$, $\mathcal{L}_{\mathbf{b}}^{\text{bce}}$, $\mathcal{L}_{\mathbf{i}}^{\text{dice}}$ and $\mathcal{L}_{\mathbf{b}}^{\text{dice}}$ denote the binary cross-entropy and dice loss for image- and batch-level mask prediction.

\paragraph{Inference.}

In CropFormer, the same entities between the full image and its crop are represented by the same queries by virtue of the association module. For the final segmentation output, we fuse the per-pixel mask predictions obtained from the full image and every crop of the 4 corners by using an average operation. The confidence score of each entity comes from the batch-level entityness prediction score.

\section{Experiments}
We conduct our main experiments on the entity segmentation with proposed EntitySeg Dataset in this work. First, we
benchmark our dataset in several segmentation tasks, including entity, instance, semantic and panoptic segmentation. Then, we ablate our proposed CropFormer to demonstrate its effectiveness.
Note that, in Sec.~\ref{sec:cropformer}, we
omit the conventional \textit{data batch dimension} to improve clarity, but in practice, we train CropFormer with a batch of multiple dataset images and their crops, following existing methods.

\noindent \textbf{Training Details.} We leave out some details like the training schedule and some ablation studies due to paper length limitations (see supplementary for details).

\subsection{Dataset Benchmarks}
\begin{table}[t!]
\centering
\scriptsize
\begin{tabular}{c|c|c|c|c|c}
Task & Metrics & Method & Backbone & Pretrain & Performance \\ \hline
\multirow{5}*{SEM} & \multirow{5}*{mIoU} & Deeplabv3 & R-50 & ImageNet & 27.9 \\ \cline{3-6}
& & \multirow{4}*{Mask2Former} & R-50 & ImageNet & 37.8 \\ 
& & & R-50 & COCO & 43.3 \\
& & & Swin-T & COCO & 45.0 \\
& & & Swin-L & COCO & 50.5 \\ \hline
\multirow{5}*{INS} & \multirow{5}*{AP} & Mask-RCNN & R-50 & ImageNet & 5.0 \\ \cline{3-6}
& & \multirow{4}*{Mask2Former} & R-50 & ImageNet & 13.0 \\
& & & R-50 & COCO & 20.3 \\
& & & Swin-T & COCO & 22.7 \\
& & & Swin-L & COCO & 30.3 \\ \hline
\multirow{5}*{PAN} & \multirow{5}*{PQ} & PanopticFPN & R-50 & ImageNet & 3.6 \\ \cline{3-6}
& & \multirow{4}*{Mask2Former} & R-50 & ImageNet & 5.5 \\
& & & R-50 & COCO & 9.6 \\
& & & Swin-T & COCO & 9.8 \\
& & & Swin-L & COCO & 13.4 \\ \hline
\multirow{5}*{ENT} & \multirow{5}*{AP$^e$} & Mask-RCNN & R-50 & ImageNet & 24.9 \\ \cline{3-6}
& & \multirow{4}*{Mask2Former} & R-50 & ImageNet & 26.0 \\
& & & R-50 & COCO & 35.2 \\
& & & Swin-T & COCO & 42.9 \\
& & & Swin-L & COCO & 46.2 \\ \hline

\end{tabular}
\caption{Benchmark on several segmentation tasks on the EntitySeg Dataset. The `SEM', `INS', `PAN', and `ENT' indicate the task of semantic, instance, panoptic and entity segmentation. We note that we leave out all details and more experiments in the supplementary file due to paper length limitations.}
\label{Tab:aba_benchmark_merge}
\end{table}

\noindent \textbf{Benchmark on EntitySeg Dataset.} In Table~\ref{Tab:aba_benchmark_merge}, we provide results on semantic, instance, panoptic, and entity segmentation on EntitySeg dataset with some baseline methods. Similar to existing segmentation datasets like COCO and ADE20K, the use of COCO pre-trained weight and a stronger backbone can benefit each segmentation performance of EntitySeg Dataset. Compared to COCO and ADE20K, EntitySeg dataset is more challenging as evidenced by the smaller performance numbers (\textit{e.g.,} Swin-L Mask2Former's PQ on COCO-PAN is 57.8\%).

\paragraph{Transfering EntitySeg to other general segmentation tasks.}
Table~\ref{Tab:aba_transfer_general_seg} shows the performance comparison of transferring models pretrained on either COCO or EntitySeg dataset to other image/video-level segmentation tasks. Owing to its emphasis on in-the-wild generalization, EntitySeg provides greater improvements than COCO dataset.
\begin{table}[t!]
\centering
\scriptsize
\setlength\tabcolsep{3pt}
\begin{tabular}{c|c|c|c|c|c}
\cellcolor{lightgray!30} Transfer Dataset &
\cellcolor{lightgray!30} Type &
\cellcolor{lightgray!30} SEM (mIoU) &
\cellcolor{lightgray!30} INS (AP) &
\cellcolor{lightgray!30} PAN (PQ) & \cellcolor{lightgray!30} ENT (AP$^e$)\\ \hline
ADE20K & Image & 57.2\textcolor[RGB]{34,139,34}{(+1.0)} & 36.0\textcolor[RGB]{34,139,34}{(+1.1)} & 
49.0\textcolor[RGB]{34,139,34}{(+0.9)} & 40.6\textcolor[RGB]{34,139,34}{(+1.5)} \\ \hline
YVIS2019~\cite{yang2019video} & Video & - & 61.3\textcolor[RGB]{34,139,34}{(+0.9)} & - & -\\
VIPSeg~\cite{miao2022large} & Video & - & - & 37.0\textcolor[RGB]{34,139,34}{(+1.1)} & - \\
UVO~\cite{wang2021unidentified} & Video & - & - & - & 29.5\textcolor[RGB]{34,139,34}{(+2.2)}\\ \hline
\end{tabular}
\caption{The relative performance gains of EntitySeg pretraining over COCO-Panoptic pretraining when fine-tuned on various image/video-level segmentation datasets.}
\label{Tab:aba_transfer_general_seg}
\end{table}

\paragraph{Benefit of EntitySeg for high-quality image segmentation.}
In Table~\ref{Tab:aba_transfer_highquality_seg}, we transfer Mask2Former pretrained on EntitySeg dataset for finetuning on Dichotomous Image Segmentation~\cite{qin2022highly}. 
Compared to state-of-the-art ISNet~\cite{qin2022highly} that is carefully designed for the task, Mask2Former pretrained on EntitySeg outperforms ISNet by a large margin, despite not having any task-specific design. This is by virtue of the high-quality mask annotations in EntitySeg dataset.
\begin{table}[t!]
\centering
\scriptsize
\setlength\tabcolsep{3pt}
\begin{tabular}{c|c|c|c|c|c|c}
\cellcolor{lightgray!30} Method &
\cellcolor{lightgray!30} $F_{\beta}^{mx} \uparrow$ &
\cellcolor{lightgray!30} $F_{\beta}^{w} \uparrow$ & \cellcolor{lightgray!30} $M \downarrow$ &
\cellcolor{lightgray!30} $S_{\alpha} \uparrow$ &
\cellcolor{lightgray!30} $E_{\phi}^{m} \uparrow$ &
\cellcolor{lightgray!30} $HCE_{\gamma} \downarrow$
\\ \hline
IS-Net~\cite{qin2022highly} & .791 & .717 & .074 & .813 & .856 & 1116 \\
Ours & \textbf{.872} & \textbf{.789} & \textbf{.035} & \textbf{.896} & \textbf{.913} & \textbf{969}\\ \hline
\end{tabular}
\caption{Evaluation on Dichotomous Image Segmentation (DIS5K dataset~\cite{qin2022highly}.) `Ours' indicates Mask2Former pretrained on EntitySeg and then finetuned on DIS5K.}
\label{Tab:aba_transfer_highquality_seg}
\end{table}

\newcommand{\myarrow}[1][1cm]{\mathrel{%
   \vcenter{\hbox{\rule[+.1pt]{#1}{.35pt}}}%
   \mkern-10mu\hbox{\usefont{U}{lasy}{m}{n}$\longrightarrow$}}}

\begin{table*}[t!]
\centering
\scriptsize
\setlength{\tabcolsep}{3pt}
\begin{tabular}{c|l|c|c|c|c|c|c}
\cellcolor{lightgray!30} Method &
\cellcolor{lightgray!30} \Gape[0pt][2pt]{\makecell{Input Preprocessing}} &
\cellcolor{lightgray!30} GPU Type &
\cellcolor{lightgray!30} AP$^e$ &
\cellcolor{lightgray!30} \makecell[c]{GPU Memory \\ (Train)} &
\cellcolor{lightgray!30} \makecell[c]{GPU Memory \\(Inference)} &
\cellcolor{lightgray!30} \makecell[c]{Train Time \\(GPU day)} &
\cellcolor{lightgray!30} \makecell[c]{Inference Time \\(ms)} \\ \hline
Mask2Former & high-res image $\stackrel{\text{resize}}{\longrightarrow}$ (800, 1333) & A100-40G & 39.5 & 5.8G & 3.0G & 2.4 & 637\\
Mask2Former & high-res image $\stackrel{\text{resize}}{\longrightarrow}$ (1040, 1732) & A100-40G & 39.8 & 6.7G & 4.1G & 4.8 & 956\\
Mask2Former & high-res image $\stackrel{\text{resize}}{\longrightarrow}$ (\textbf{2700}, \textbf{4500}) & A100-40G & OOM & OOM & - & OOM & - \\
Mask2Former & high-res image $\stackrel{\text{resize}}{\longrightarrow}$ (\textbf{2700}, \textbf{4500}) & \textbf{A100-80G} & 40.6 & \textbf{58.8G} & \textbf{10.4G} & \textbf{60.8} & \textbf{3312} \\
MS-Mask2Former & high-res image $\stackrel{\text{resize}}{\longrightarrow}$  (800:1120, 1732) & A100-40G & 39.8 & 8.7G & 6.4G & 2.8 & 2594 \\ \hline 
V-Mask2Former~\cite{cheng2021mask2former} & \multirow{3}{*}{high-res image \begin{tabular}{@{}c@{}} $\stackrel{\text{resize}}{\myarrow[59px]}$ (800, 1333) \\ \hspace{-2px}$\underset{\text{crop(}\delta\text{)}}{\longrightarrow}$ high-res crop $\underset{\text{resize}}{\longrightarrow}$ (800, 1333) \end{tabular}} & A100-40G & 39.7 & 10.4G & 6.7G & 6.8 & 1503\\
VITA~\cite{heo2022vita} & & A100-40G & 39.9 & 11.9G & 6.8G & 10.2 & 1512\\ \cline{1-1}
\cline{3-8}
CropFormer  & &  A100-40G &\textbf{41.0} & 11.4G & 6.9G & 9.8 & 1503 \\ \hline
\end{tabular}
\caption{Comparison with high-resolution methods on EntitySeg with multi-scale training. `(x, y)' indicates number of pixels along the shortest (x) and longest (y) sides of image. For multi-scale (MS) Mask2Former, `x1:x2' randomly samples from the range of $[\text{x1}, \text{x2}]$ with a step size of 32 during training. Row 1, 2, \& 3 are single-scale Mask2Formers trained and evaluated at various single-scale resolutions. The remaining rows: (MS-Mask2Former) during inference, Hungarian matching is used to naively ensemble MS-Mask2Former's multi-scale results, by traversing all scales defined by $[\text{x1}, \text{x2}]$; (V-Mask2Former \& VITA) mask predictions from full image and four crops are fused with either video-level Mask2Former or VITA.}
\label{Tab:aba_cropformer_cost}
\end{table*}
\paragraph{Segmentation `in the wild'}
Table~\ref{Tab:aba_transfer_wild_seg} compares in-the-wild segmentation performance between Mask2Formers trained on either EntitySeg or COCO. Here, we directly use the trained models without finetuning on the evaluated datasets. The EntitySeg outperforms COCO-Panoptic \cite{kirillov2019panoptic} and COCO-Entity \cite{qi2021open} by large margins.

\begin{table}[t!]
\centering
\scriptsize
\setlength\tabcolsep{3pt}
\begin{tabular}{c|c|c|c|c|c}
\cellcolor{lightgray!30} Train Dataset & 
\cellcolor{lightgray!30} Task & 
\cellcolor{lightgray!30} LVIS~\cite{gupta2019lvis} &
\cellcolor{lightgray!30} OCID~\cite{DBLP:conf/icra/SuchiPFV19} &
\cellcolor{lightgray!30} CAMO~\cite{le2021camouflaged} &
\cellcolor{lightgray!30} FSS~\cite{li2020fss} \\ \hline
COCO & PAN & 24.38 & 39.94 & 43.80 & 64.96 \\
COCO & ENT & 27.75 & 45.88 & 59.70 & 75.60 \\ \hline
EntitySeg & ENT & \textbf{33.56} & \textbf{76.63} & \textbf{73.20} & \textbf{86.26} \\ \hline
\end{tabular}
\caption{In-the-wild segmentation performance (AR@100 with IoU threshold 0.5). We directly evaluate models trained on either EntitySeg or COCO on large-vocabulary segmentation LVIS~\cite{gupta2019lvis}, object clutter indoor segmentation for robot grab~\cite{DBLP:conf/icra/SuchiPFV19}, camouflaged instance segmentation CAMO~\cite{le2021camouflaged}, few-shot segmentation FSS~\cite{li2020fss}. All the models are based on Mask2Former with Swin-Large backbone.}
\label{Tab:aba_transfer_wild_seg}
\end{table}

\subsection{A Comprehensive Analysis of CropFormer}
We first ablate Cropformer on the entity segmentation task and then report the results of CropFormer on other segmentation datasets.

Table~\ref{Tab:aba_cropformer_cost} shows the benefits of CropFormer on segmentation performance and computation cost compared to the Mask2Former~\cite{cheng2022masked}, its multi-scale extension, as well as related methods
~\cite{cheng2021mask2former,heo2022vita} that supports multi-view fusion. The first four rows are our baseline Mask2Former with the single-scale inference (800, 1040, or 2700 for shortest side) and (1333, 1732 or 4500 for longest side) for the full images, where 1040=800$\times\delta$ and 1732=1333$\times\delta$ with $\delta=0.7$ by default. For the results in the fifth row, we use test-time Hungarian matching to associate the same entities obtained with multi-scale images. We find 
minimal improvement with such an inference strategy. In the sixth and seventh rows, directly using video-level Mask2Former or VITA, the state-of-the-art video instance segmentation framework, merely brings marginal performance gains. Whereas, using the crop output from CropFormer's batch decoder achieves a significant AP$^\text{e}$ gain as indicated by the last row. 
By combining the full image and four crop outputs from CropFormer (final row), we obtain even stronger 1.5 AP$^\text{e}$ gains compared to the baseline (first row). Compared to the Mask2Former baselines with a naive increase in image resolution, CropFormer is far more efficient computationally.


\begin{table}[t!]
\begin{minipage}{\textwidth}
\begin{minipage}[t]{0.16\textwidth}
            \centering
            \scriptsize
            \setlength{\tabcolsep}{2pt}
            \begin{tabular}{c|c|c}
            \cellcolor{lightgray!30} Iterations (Train) & \cellcolor{lightgray!30} Mask2Former &  \cellcolor{lightgray!30} CropFormer  \\  \hline
            1$\times$ & 39.5 & 41.0 \textcolor[RGB]{34,139,34}{(+1.5)}  \\ 
            2$\times$ & 40.2 & 42.1 \textcolor[RGB]{34,139,34}{(+1.9)}  \\ 
            3$\times$ & 40.9 & 42.8 \textcolor[RGB]{34,139,34}{(+1.9)}\\ 
            4$\times$ & 40.9 & 42.7 \textcolor[RGB]{34,139,34}{(+1.8)} \\ \hline
            \end{tabular}
        
        (a) metric: AP$^{\text{e}}$
        \end{minipage}
        \begin{minipage}[t]{0.46\textwidth}
        \centering
        \scriptsize
        \setlength{\tabcolsep}{2pt}
        \begin{tabular}{c|ccc}
        \cellcolor{lightgray!30} Decoder & \cellcolor{lightgray!30} AP$^{\text{e}}$ & \cellcolor{lightgray!30} AP$^{\text{e}}_{50}$ & \cellcolor{lightgray!30} AP$^{\text{e}}_{75}$ \\ \hline
        I-O & 39.3 & 56.7 & 39.8 \\ 
        B-O & 39.1 & 56.6 & 39.7 \\ 
        B-C & 40.2 & 57.5 & 40.8 \\
        B-OC & \textbf{41.0} & \textbf{58.4} & \textbf{41.9} \\ \hline
        \end{tabular}
        
    (b)
        \end{minipage}

\end{minipage}
\caption{Ablation study on: (a) training schedules; (b) CropFormer's decoder combinations. In sub-table(b), the `Decoder' column indicates whether we use the predictions of the full image (`O'), four cropped patches (`C'), or both of them (`OC') from the `Image' or `Batch' decoder.}
\label{Tab:aba_iteration_and_fusion}
\end{table}


\paragraph{Longer Training Schedules} 
Table~\ref{Tab:aba_iteration_and_fusion}(a) indicates that CropFormer consistently improves over the baseline Mask2Former when we adopt 2$\times$/3$\times$ 
training schedules. A similar trend is observed when using Swin-L~\cite{liu2021swin} backbone, which is provided in our supplementary file.
\paragraph{Decoder Performance} Table~\ref{Tab:aba_iteration_and_fusion}(b) shows the impact of using different decoder combinations in CropFormer for mask prediction during inference. Using just the crop outputs from CropFormer's batch decoder (B-C) outperforms using just the full image view (I-O, B-O). Combining both the full image and crops (B-OC) gives the best result. 
\paragraph{CropDataloader} Table~\ref{abl_1}(a) and (b) show the ablation study on the usage of crop ratio and crop type. As indicated by Table~\ref{abl_1}(a), CropFormer obtains the best performance with a crop ratio $\delta$ of 0.7. Smaller crop ratios perform worse and it might be a problem caused by queries that are not robust against drastic image transformation, which we leave for future work. Table~\ref{abl_1}(b) shows the impacts of the type of crop and the number of crops. 4 and 8 fixed crops during training and testing perform the best.
\paragraph{Association Module} Table~\ref{abl_1}(c) shows the ablation study on the structure of the association module. There is a slight performance difference between using self-attention or a feedforward module following cross-attention.

\begin{table}[t!]
\begin{minipage}{\textwidth}
\begin{minipage}[t]{0.08\textwidth}
            \centering
            \scriptsize
             \setlength{\tabcolsep}{2pt}
            \begin{tabular}{c|c}
            \cellcolor{lightgray!30} $\delta$ & \cellcolor{lightgray!30} AP$^\text{e}$ \\ \hline
            0.5 & 38.5 \\ 
            0.6 & 40.2  \\ 
            0.7 & \textbf{41.0}  \\ 
            0.8 & 40.9 \\ \hline
            \end{tabular}
        
        (a)
        \end{minipage}
        \begin{minipage}[t]{0.22\textwidth}
        \centering
        \scriptsize
        \setlength{\tabcolsep}{2pt}
        \begin{tabular}{c|c|c}
        
            \cellcolor{lightgray!30} Train & \cellcolor{lightgray!30} Test & \cellcolor{lightgray!30} AP$^\text{e}$ \\ \hline
            Random & Fixed (4) & 39.7\\ 
            Fixed (4) & Fixed (4)  & 41.0  \\
            Fixed (4) & Fixed (8)  & \textbf{41.3} \\ 
            Fixed (8) & Fixed (8)  & 41.0 \\ \hline
        \end{tabular}
        
    (b)
        \end{minipage}
        \begin{minipage}[t]{0.15\textwidth}
        \centering
        \scriptsize
        \setlength{\tabcolsep}{2pt}
        \begin{tabular}{ccc|c}
            \cellcolor{lightgray!30} XAtt & \cellcolor{lightgray!30} SAtt & \cellcolor{lightgray!30} FFN & \cellcolor{lightgray!30} AP$^\text{e}$ \\ \hline
            \checkmark & $\circ$ & $\circ$ &  40.7 \\ 
            \checkmark & $\circ$ & \checkmark & 40.8  \\ 
            \checkmark & \checkmark & $\circ$ & 40.8 \\
            \checkmark & \checkmark & \checkmark & \textbf{41.0} \\ \hline
        \end{tabular}
        
    (c)
    \end{minipage}

\end{minipage}
\caption{Ablation study on the usage of crop ratio $\delta$, crop type and association module in CropFormer. In sub-table (b), `Random' indicates random crops and `Fixed (4/8)' indicates 4 or 8 fixed corner crops. In sub-table(c), \checkmark and $\circ$ means whether we use the module or not.}
\label{abl_1}
\end{table}

\begin{table}[t!]
\begin{minipage}{\textwidth}
\begin{minipage}[t]{0.22\textwidth}
        \centering
        \scriptsize
        \setlength{\tabcolsep}{2pt}
        \begin{tabular}{c|c|c}
            \cellcolor{lightgray!30} Method & \cellcolor{lightgray!30} AP$^e$ & \cellcolor{lightgray!30} AP$^b$ \\ \hline
            Mask Transfiner~\cite{ke2022mask} & 33.7 & 25.0 \\
            PatchDCT~\cite{wen2023patchdct} & 35.4 & 27.4 \\ \hline 
            Mask2Former & 39.5 & 31.2 \\ 
            Ours & \textbf{41.0} & \textbf{33.7} \\ \hline
        \end{tabular}
        
    (a)
        \end{minipage}
        \begin{minipage}[t]{0.23\textwidth}
        \centering
        \scriptsize
        \setlength{\tabcolsep}{2pt}
        \begin{tabular}{c|c|c} \cellcolor{lightgray!30} Task & \cellcolor{lightgray!30} Mask2Former & \cellcolor{lightgray!30} CropFormer \\ \hline
            COCO-INS & 50.1 & \textbf{51.9} \\ 
            COCO-PAN & 57.8 & \textbf{59.1} \\ 
            ADE20K-INS & 34.9 & \textbf{36.4} \\
            ADE20K-PAN & 48.1 & \textbf{49.7} \\ \hline
        \end{tabular}
        
    (b) metrics: AP for INS; PQ for PAN
    \end{minipage}
\end{minipage}
\caption{
(a) Comparison with high-quality segmentation methods on 
EntitySeg task/dataset using AP$^e$ and boundary AP (AP$^b$). (b) evaluation on instance (INS) and panoptic (PAN) segmentation tasks on COCO and ADE20K datasets.}
\label{abl_2}
\end{table}

\paragraph{Comparison to other high-quality segmentation methods.} Table~\ref{abl_2}(a) shows that our CropFormer 
significantly outperforms recent high-quality instance-level segmentation methods like Mask Transfiner~\cite{ke2022mask} and PatchDCT~\cite{wen2023patchdct} on both traditional and boundary APs. For a fair comparison, all the methods in Table~\ref{abl_2}(a) use Swin-Tiny backbone and COCO pre-trained weights.
\paragraph{Improvement on class-aware segmentation tasks} Table~\ref{abl_2}(b) shows the 
class-aware task performance of CropFromer with Swin-L~\cite{liu2021swin} backbone and 3$\times$ training schedule. CropFormer can still achieve competitive gains on the instance and panoptic segmentation tasks on COCO and ADE20K datasets. This suggests that CropFormer may provide other benefits that are beyond high-resolution segmentation (\textit{e.g.,} prediction robustness).

\section{Conclusion}
This paper proposes an EntitySeg dataset and a CropFormer framework for high-quality entity segmentation, with a strong focus on in-the-wild generalization and high-quality dense segmentation. The EntitySeg dataset contains about 33K images from various domains and resolutions, accompanied by high-quality mask annotations for training and testing. A novel CropFormer framework is also proposed to 
overcome the computation challenges introduced by
EntitySeg dataset's high-quality and high-resolution characteristics. Specifically, it enables fusing multi-view predictions from a low-res full image and its high-res crops by learning batch-level queries. We hope this work will serve as a catalyst for future research on open-world and high-quality image segmentation tasks.
{\small
\bibliographystyle{ieee_fullname}
\bibliography{egbib}
}

\end{document}